\documentclass[twoside,leqno,twocolumn]{article}
\usepackage{ltexpprt}
\usepackage{amsmath}
\usepackage{amssymb}
\usepackage{graphicx}
\usepackage{array}
\usepackage{algorithm}
\usepackage[noend]{algorithmic}
\usepackage[sort]{cite}

\DeclareMathOperator*{\argmax}{arg\,max}

\begin{document}

\title{\Large Finding Competitive Network Architectures Within a Day Using UCT}
\author{Martin Wistuba\thanks{martin.wistuba@ibm.com, IBM Research AI -- Ireland.}}
\date{}

\maketitle







\begin{abstract} \small\baselineskip=9pt
The design of neural network architectures for a new data set is a laborious task which requires human deep learning expertise.
In order to make deep learning available for a broader audience, automated methods for finding a neural network architecture are vital.
Recently proposed methods can already achieve human expert level performances.
However, these methods have run times of months or even years of GPU computing time, ignoring hardware constraints as faced by many researchers and companies.
We propose the use of Monte Carlo planning in combination with two different UCT (upper confidence bound applied to trees) derivations to search for network architectures.
We adapt the UCT algorithm to the needs of network architecture search by proposing two ways of sharing information between different branches of the search tree.
In an empirical study we are able to demonstrate that this method is able to find competitive networks for MNIST, SVHN and CIFAR-10 in just a single GPU day.
Extending the search time to five GPU days, we are able to outperform human architectures and our competitors which consider the same types of layers.
\end{abstract}

\section{Introduction.}
Deep learning methods are very successful in various applications such as machine translation \cite{Sutskever2014}, image and speech recognition \cite{He2016,Szegedy2015,Dahl2012} or reinforcement learning in general \cite{Arulkumaran2017}.
The reason for its success is the ability to learn how to extract features from unstructured data.
Hence, we observe a shift from the laborious manual feature engineering task for audio, images and text to engineering network components and architectures.
While achieving overall better performances, this still involves a laborious, manual task which requires experts.
Thus, true end-to-end learning remains an important and active topic of research.

Earliest work is based on neuro-evolution \cite{Stanley2002}.
Mutations and cross-over operations are used to adapt the network structure and even learn the network weights.
This is a very computational expensive procedure.
However, constraining the genetic algorithm to use only mutations and learning the parameters with gradient-based optimization methods, it is possible to discover good network architectures if enough computational power is available \cite{Real2017}.

Recently, various approaches using reinforcement learning have been proposed.
Some approaches are based on recurrent neural networks, Q-Learning or Monte Carlo planning which learn to select layer by layer \cite{Baker2017, Negrinho2017, Zoph2017}.
Others try to learn how to change an existing architecture to improve the results \cite{Cai2017}.
Yet, most of these approaches require a vast amount of computational power.

In this work we propose a method that is computational feasible and still finds competitive networks.
We define the search problem as a Markov decision problem where the state-action graph is a tree.
We propose to search for network architectures by maximizing the expected reward using Monte Carlo planning.
In particular, we derive two different policies based on UCT \cite{Kocsis2006} and compare our method to the state-of-the-art in an empirical evaluation on MNIST, SVHN and CIFAR-10.
We show in our empirical evaluation that our proposed methods can find competitive network architectures in just one GPU day with a minimal loss in accuracy.

\section{Related Work.}
Algorithm selection and hyperparameter optimization are a very old problem \cite{Rice1976} and not limited to machine learning applications.
Bayesian optimization \cite{Snoek2012} is often considered to be the state-of-the-art for automated hyperparameter optimization.
One of its drawbacks is that it expects a fixed length encoding of the algorithm and hyperparameter choice and that the validation score function is smooth with respect to this encoding.
While Bayesian optimization was able to achieve state-of-the-art results for a fixed architecture \cite{Snoek2012,Snoek2015}, no encoding for network architectures has been found which satisfies this assumption.
Nonetheless, some works apply Bayesian optimization to search for network architectures \cite{Swersky2014,Mendoza2016}.
Recently, various works have shown that reinforcement learning and neuro-evolution can find neural networks which achieve results similar to human-engineered architectures \cite{Cai2017,Baker2017,Zoph2017,Real2017}.

The idea of using genetic algorithms in order to adapt neural networks is a relatively old idea.
Early work only evolved the weights of a fixed architecture \cite{Miller1989}, later work also evolved the architecture \cite{Stanley2002}.
A recent work based on neuro-evolution was able to find competitive convolutional neural networks for CIFAR-10 \cite{Zoph2017}.
The authors propose to adapt the network architecture with random mutations.
They do not use any cross-over operations and estimate the network weights using gradient-based optimization methods.
However, this optimization method remains computational very expensive.

Several optimization methods based on reinforcement learning have been proposed for the neural network architecture search.
Cai et al. propose to learn a policy that is able to improve an existing network by deepening or widening it \cite{Cai2017}.
Others address the task to find architectures from scratch \cite{Baker2017,Real2017}.
They propose to learn a policy which estimates the final architecture by choosing layer by layer.
These approaches have in common with neuro-evolution that they are computational expensive.

Negrinho and Gordon \cite{Negrinho2017} investigate various optimization methods for our problem, among them UCT.
In order to make UCT feasible, they propose to combine it with a bisection method to share information among similar actions in the same state.
In contrast to them, we propose two different variants of UCT.
One of them shares information for the same action in similar states and the other is sharing information between similar actions by predicting the final reward based on previously chosen actions.

Our main difference to the current work is that we are aiming at finding a competitive network architecture within a time frame that is affordable for all researchers.
Hence, we will give our method the symbolic time budget of just a single GPU day.

\section{Problem Definition.}\label{sec:problem_definition}
We define the search for neural network architectures as a Markov decision problem described by the tuple $\left(S,A,P,R,\gamma\right)$.
Each state of the state space $S$ describes the current network architecture and each action in the action set $A$ adds another layer.
Thus, we can describe a state by the actions taken so far and the state set $S$ is defined as a subset of all action permutations, $S\subset\mathcal{P}\left(A\right)$.
The initial state is the empty set, $S_{0}=\emptyset$, and final states contain a termination action.
In order to allow a fair comparison to the work by Baker et al. \cite{Baker2017}, we are defining an action space that is searching in a very similar network architecture search space.

We define following 19 actions: twelve different actions that are adding convolutional layers with square kernel sizes 1, 3 or 5 with 64, 128, 256 or 512 filters each with stride 1 and same zero padding.
Three pooling actions adding layers with square pooling sizes 2, 3 and 5 with strides 2, 2 and 3, respectively.
Another three actions adding fully connected layers with 128, 256 or 512 units and finally one termination action which adds a softmax layer.

Not every action can be chosen in every state in order to obtain legitimate network structures.
As soon as the termination action was chosen, a final state is reached and no further actions are possible.
Actions adding convolutional layers are possible at any point before the first addition of a fully connected layer.
Pooling actions are only allowed immediately after a convolutional action.
The first pooling action can be chosen as soon as the first convolutional action with kernel size greater than one has been selected.
For computational reasons, the first fully connected layer can be selected as soon as the input dimension of the feature map is smaller than eight.
There can be at most two consecutive fully connected actions whereas the second fully connected layer can have at most as many units as the first one.

This Markov decision process is fully deterministic and hence the state transition probabilities $P$ are either zero or one.
We define the reward function as
\begin{equation}
R_{a}\left(s,s'\right)=\begin{cases}
\operatorname{acc}\left(s'\right) & \text{if }a\text{ is a terminal actions.}\\
0 & \text{otherwise}
\end{cases}\enspace,
\end{equation}
where $\operatorname{acc}\left(s'\right)$ is the accuracy obtained when training the network described by state $s'$ on the training split and evaluating it on the validation split.

We set the discount factor $\gamma=1$ because we do not want to punish deeper architectures.

Our objective is to find the network architecture with maximal accuracy on the validation split after training on the training split,
\begin{equation}
\mathcal{O}\left(S\right)=\argmax_{s\in S}\operatorname{acc}\left(s\right)\enspace.
\end{equation}
To achieve this, we present in the upcoming section various methods which maximize
\begin{equation}
\mathcal{O}\left(S\right)=\argmax_{\pi}\sum_{t=0}^{\infty}\gamma^{t}R_{\pi\left(s_{t}\right)}\left(s_{t},s_{t+1}\right)
\end{equation}
as a proxy objective by using an adaptive policy
\begin{equation}
\pi\ :\ S\rightarrow A\enspace.
\end{equation}

\section{Architecture Search with Monte Carlo Planning.}

In this section we are briefly reviewing Monte Carlo planning and the UCT algorithm.
Then, we show how these methods can be applied for the task of finding competitive neural network architectures and describe necessary adaptions and our improvements.
Finally, we describe a transfer learning method which we are using in order to speed up the search.

\subsection{Monte Carlo Planning.}
Monte Carlo planning is one approach to find optimal policies in large Markov decision problems \cite{Silver2010}.
The most famous work in this domain is the UCT algorithm \cite{Kocsis2006} which led to big advances in artificial intelligence for many games, in particular the game of Go \cite{Gelly2006,Gelly2007}.
The UCT algorithm is designed for state-action spaces which have a tree structure.
It is a rollout-based algorithm that works as follows.
A part of the state-action tree is stored, starting with the root only.
In each rollout the algorithm follows the tree policy as long as the current state is within the stored tree.
The tree policy is based on the UCB1 formula \cite{Auer2002}
\begin{equation}
 \pi_{\text{tree}}\left(s\right)=\argmax_{a\in A}\frac{R_{s,a}}{n_{s,a}}+c\sqrt{\frac{\ln\left(n_{a',s}\right)}{n_{s,a}}}\enspace,\label{eq:uct}
\end{equation}
where $R_{s,a}$ is the cumulative reward received when selecting action $a$ in state $s$, $n_{s,a}$ is the number of times action $a$ was chosen when
being in state $s$ and $n_{a',s}$ is the total number of times an action $a'$ was considered that leads to state $s$ in a rollout.
The constant $c$ controls the trade-off between exploration and exploitation.

As soon as a new state is reached, the tree is expanded by this node.
Actions are now chosen uniformly at random until a leaf node is reached.
The reward is estimated and backpropagated.

The UCT algorithm has provable regret bounds and converges against the optimal policy given enough time \cite{Kocsis2006}.
Furthermore, it is trivially parallelizable and can be stopped and continued at any time.

\subsection{UCT for Architecture Search.}\label{sub:rave4nn}
Since the state-action space as described in Section \ref{sec:problem_definition} is a tree, the UCT algorithm can in principle be applied to our problem without any changes.
However, this search space is very large and one rollout is very time-consuming since it involves training and testing a neural network.
We can only afford few hundreds of rollouts but then we spend most of the time exploring the search space.
The reason for this is that according to the UCB1 formula each action in a state has to be chosen at least once before an action is considered a second time.
This is a general problem for UCT during the first rollouts.
Thus, Gelly et al. \cite{Gelly2007} proposed the rapid action value estimation (RAVE) for UCT.
They make the assumption that it does not matter when an action is selected but if.
This allows sharing information between different branches of the search tree but is limited to domains where action sequences can be transposed.
Obviously, this is not the case for neural network architectures since the order of layers is important.
Thus, we propose an alternative way of sharing information between different tree branches.
We combine the information for actions happening at the same layer depth and in similar states.
We define the rapid action value estimation for neural networks policy (\textsc{RAVE4NN}) as
\begin{eqnarray}
\pi_{\textsc{RAVE4NN}}\left(s\right) & = & \argmax_{a\in A}\left(\alpha\frac{R_{s,a}}{n_{s,a}}+\beta\frac{R_{\mathcal{N}\left(s\right),a}}{n_{\mathcal{N}\left(s\right),a}}\right)\nonumber\\
 & + & c\sqrt{\frac{\ln\left(\alpha\cdot n_{a',s}+\beta\cdot n_{a',\mathcal{N}\left(s\right)}\right)}{\alpha\cdot n_{s,a}+\beta\cdot n_{\mathcal{N}\left(s\right),a}}}\label{eq:rave4nn}
\end{eqnarray}
where $\mathcal{N}\left(s\right)$ is the set of states being similar to state $s$ and $\alpha$ and $\beta$ are abbreviations for the weighting functions
$\alpha\left(n_{s,a},n_{\mathcal{N}\left(s\right),a}\right)$ and $\beta\left(n_{s,a},n_{\mathcal{N}\left(s\right),a}\right)$, respectively.
Using the notation from Equation \eqref{eq:uct},
\begin{eqnarray}
R_{\mathcal{N}\left(s\right),a} & = & \sum_{s'\in\mathcal{N}\left(s\right)}R_{s',a}\\
n_{\mathcal{N}\left(s\right),a} & = & \sum_{s'\in\mathcal{N}\left(s\right)}n_{s',a}\\
n_{a',\mathcal{N}\left(s\right)} & = & \sum_{s'\in\mathcal{N}\left(s\right)}n_{a',s'}\enspace.
\end{eqnarray}

Given are two state-action sequences $\left(s_{0},a_{1,1},s_{1,1},a_{1,2},\ldots,s_{1,t_{1}},\ldots\right)$ and $\left(s_{0},a_{2,1},s_{2,1},a_{2,2},\ldots,s_{2,t_{2}},\ldots\right)$.
We define the similarity between two states as follows.
If two states $s_{1,t_{1}}$ and $s_{2,t_{2}}$ are similar, $s_{1,t_{1}}\in\mathcal{N}\left(s_{2,t_{2}}\right)$, then they are representing a network part of the same depth, i.e. $t_{1}=t_{2}$.
Furthermore, the output dimension of this network part belongs to the same representation bin.
We distinguish three bins with the representation size being in the interval $\left[1,4\right)$, $\left[4,8\right)$ or $\left[8,\infty\right)$.
Finally, the number of chosen fully connected actions is equal before reaching the states $s_{1,t_{1}}$ and $s_{2,t_{2}}$.
Only for softmax actions it is additionally required that the action before to the current state is the same, $a_{t_{1}}=a_{t_{2}}$.

\subsection{Contextual Reward Predictions.}\label{sub:crp}

In this section we propose to share information among search tree branches by means of a prediction model.
The prediction model is forecasting the reward for taking an action given a state.
In contrast to the approach in the previous section, this enables us to predict rewards even in cases where RAVE4NN cannot infer anything from similar cases.
Furthermore, this also enables us to learn from data sets investigated in previous experiments for a new data set and learn policies across problems.
However, in this work we do not further investigate this possibility.

We define the contextual reward prediction policy as
\begin{equation}
 \pi_{\text{CRP}}\left(s\right)=\argmax_{a\in A}\hat{R}\left(s,a\right)+c\sqrt{\frac{\ln\left(n_{d}\right)}{n_{d,a}}}\label{eq:crp}\enspace.
\end{equation}
$\hat{R}$ is a prediction model predicting the reward based on the action and the state.
$n_{d}$ is the number of times a state with $d$ layers was reached and $n_{d,a}$ the number of times action $a$ was chosen at depth $d$.
We encode the state $s'$, reached by taking action $a$ in state $s$, by representing a network of depth $d$ with a vector $\mathbf{x}\in\mathbb{R}^{d+2}$.
The first $d$ entries of this vector contain the log number of parameters per layer.
There are two more entries for the log number of total parameters and the log representation size.
Predictors for convolutional actions have another entry for the number of filters defined by action $a$.
There exists one of these predictors for each depth and for each action.
Labeled examples are collected during the search.
For convolutional actions, we share examples among actions with the same kernel size to allow the predictors to learn across different number of filters.
We use a Gaussian process with Mat\'{e}rn 5/2 kernel as a prediction model.

\subsection{Gaining Speed with Net2Net.}
Obviously, the more rollouts can be conducted, the closer we are to the optimal policy.
In order to speed up the rollouts, we are using Net2Net \cite{Chen2016}.
Net2Net is a knowledge transfer approach which widens layers, increases kernel sizes and deepens networks without changing its predictions.
The changed network will converge in less epochs than the same network initialized with random weights.
During our search we store all networks with parameters and whenever a new network needs to be initialized, we first compute the network edit distance to the networks already investigated.
We define this network edit distance by
\begin{eqnarray*}
d_{i,j} & = & \begin{cases}
d_{i-1,j-1} & \text{if }a_{i}=a_{j}\\
\min\left\{ \begin{array}{c}
d_{i,j-1}+c_{\text{ins}}\left(a_{i},a_{j}\right)\\
d_{i-1,j-1}+c_{\text{sub}}\left(a_{i},a_{j}\right)
\end{array}\right\}  & \text{otherwise}
\end{cases}
\end{eqnarray*}
where we define the costs for inserting or substituting a layer by
\begin{eqnarray*}
c_{\text{ins}}\left(a_{i},a_{j}\right) & = & \begin{cases}
\infty & \text{if }a_{j}\text{ is pooling layer or has}\\
& \text{less filters/units than }a_{i}\\
1 & \text{if }a_{i}\text{ and }a_{j}\text{have the same}\\
& \text{number of filters/units}\\
2 & \text{otherwise}
\end{cases}
\end{eqnarray*}
and
\begin{eqnarray*}
c_{\text{sub}}\left(a_{i},a_{j}\right) & = & \begin{cases}
\infty & \text{if }a_{i}\text{ or }a_{j}\text{ is pooling layer or }\\
& a_{j}\text{ has less filters/units than }a_{i}\\
2 & \text{if }a_{j}\text{ has a larger kernel size}\\
& \text{and a larger number of filters}\\
& \text{than }a_{i}\\
1 & \text{otherwise}\enspace.
\end{cases}
\end{eqnarray*}
Not every edit operation is supported by Net2Net such that the costs for an operation is infinity under certain circumstances.
This distance can be efficiently computed with dynamic programming.
We consider networks only with a network edit distance of up to two.
If no candidate can be found, we initialize the parameters at random.

\section{Experimental Results.}

In this section we summarize the results of our empirical study.
We first describe our direct competitor methods, then the data sets we used for evaluation as well as our post-processing.
We provide insights into how our proposed policies RAVE4NN (Section \ref{sub:rave4nn}) and CRP (Section \ref{sub:crp}) develop over time.
Furthermore, we compare our found network architecture to human proposed network architectures which are also using only convolutions and pooling layers.
In our supplementary material we provide an additional experiment where we compare directly to human performance on the recently released Fashion-MNIST data set.

\subsection{Competitor Methods.}
We compare our approach to five recently proposed competitor methods.
\subsubsection{Neural Architecture Search.}
Zoph and Le \cite{Zoph2017} propose to learn a policy using a recurrent neural network.
Trained with the REINFORCE rule by Williams \cite{Williams1992}, the recurrent neural network is recommending the number of filters and kernel size of a convolution layer by layer.
Furthermore, it is predicting skip connections \cite{He2016} between layers.
Only convolutional layers are considered.
After training 12,800 different architectures, trained concurrently on 800 GPUs, they found the best network architecture.
A final grid search optimizes the learning rate, weight decay, batchnorm epsilon and at what epoch to decay the learning rate.
We report the results achieved with their variant which uses the least human intervention during the optimization.

\subsubsection{Large-Scale Evolution.}
Real et al. \cite{Real2017} propose to use an evolutionary algorithm in order to search for neural network architectures.
Starting with the simplest network, it keeps improving by mutation and selection methods.
Selection is achieved by selecting two individua at random and keeping the better performing one.
A mutation is selected uniformly at random to generate a child.
Mutations can alter the learning rate, add or remove convolutional layers or skip connections and alter other properties such as the kernel size or number of filters.
Similar to Neural Architecture Search, only convolutional layers are considered.
Their final search is less computational expensive, using 250 GPUs for more than 11 days.

\subsubsection{MetaQNN.}
Baker et al. \cite{Baker2017} propose to learn a policy using Q-learning with $\epsilon$-greedy strategy and experience replay in order to learn how to create a convolutional neural network layer by layer.
In order to allow a fair comparison, we selected the same action space as we defined in Section \ref{sec:problem_definition}.

\subsubsection{Reinforcement Learning with Neural Network Transformation.}
The reinforcement learning approach by Cai et al. \cite{Cai2017} starts from an existing network and learns a policy that learns how to widen and deepen networks.
They start with an eight layer convolutional neural network with a reasonable architecture.
While the authors argue that their network is performing poor, the main reason is that there are too few parameters per layer.
Only widening the starting layers will result into a well performing network already.
Similar to us, they use Net2Net knowledge transfer to improve the speed.
In contrast to our setup, they can always profit from this knowledge transfer.

\subsubsection{DeepArchitect.}
Negrinho and Gorden \cite{Negrinho2017} investigate how the performance of Random Search \cite{Bergstra2012}, Bayesian Optimization \cite{Snoek2012} and UCT compares for optimizing the neural network architecture.
They also recognize that using vanilla UCT will not work for this task due to the limitations discussed above.
Therefore, they propose to combine UCT with bisection.
This method will trade the width of the search tree for depth and basically shares knowledge of similar actions.
However, they have a very strict definition of similar actions, requiring that the network before two similar actions is identical.
In conclusion, this approach still spends most of its time exploring the state space.
In our experiments we denote with DeepArchitect the results obtained with UCT with bisection which is the best performing approach in their paper.

\begin{figure*}[t]
 \begin{center}
 \includegraphics[width=\textwidth]{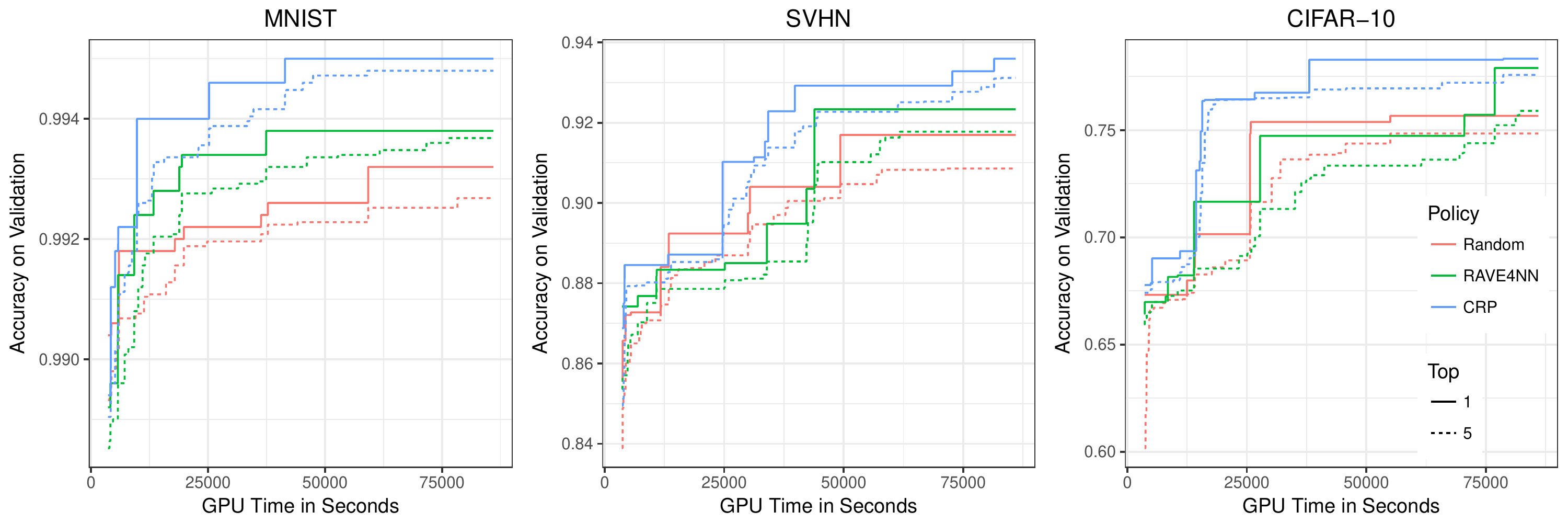}
 \caption{\label{fig:search-progress}
  The search progress for the three different policies on MNIST, SVHN and CIFAR-10.
  CRP is clearly outperforming the other policies.
  We report the validation performance of the best network so far and the mean performance of the top five networks.
  }
  \end{center}
\end{figure*}
\subsection{Implementation Details.}
Our Monte Carlo planning approach uses following hyperparameters independently on the investigated data set.
We set the exploration-exploitation trade-off in Equation \eqref{eq:uct} to $c=0.5$.
To make better use of the Net2Net knowledge transfer, we start with a maximum allowed network depth of three and increase it by one every 50 rollouts.
In case we revisit a network architecture we evaluated already, we do not train it again but use the previously computed classification accuracy.
As proposed by Baker et al. \cite{Baker2017}, dropout layers are added after every second layer and each fully connected layer with linear increasing dropout rate to 0.5.
We set the batch size to 128 and used the Adam optimizer \cite{Kingma2014} with an initial learning rate 0.001, $\beta_{1}=0.9$, $\beta_{2}=0.999$, $\epsilon=10^{-8}$ and no decay.
If the model is not significantly better than a random predictor after the first epoch, we decrease the learning rate by a factor of 0.4.
We repeat this at most five times.
If it is possible to initialize with Net2Net knowledge transfer, we train for just a single epoch.
Otherwise, we use the Glorot uniform initialization \cite{Glorot2010} and train for five epochs.

\subsection{Post-Processing.}\label{sub:post_processing}
We searched for architectures on the following image classification data sets with the setup described above.
Since only five epochs of training are not sufficient, we apply a similar, however simpler, post-processing as done by Baker et al. \cite{Baker2017}.
For example, we are not optimizing the hyperparameters but select those as chosen by Baker et al. instead.
The network with the best score on the validation is selected, reinitialized and trained according to following policies.

\subsubsection{MNIST.}
The MNIST data set \cite{Lecun1998} contains ten different classes, 60,000 training and 10,000 testing $28\times 28$ grayscale images.
The task is to identify handwritten digits.
We subtracted the global mean of each image and trained the final model for 40 epochs using the Adam optimizer settings described above.
However, we decrease the learning rate by a factor of 0.2 every five epochs.

\subsubsection{Street View House Numbers (SVHN).}
The task of the SVHN data set \cite{Netzer2011} is similar to the MNIST data set, i.e. identifying digits.
However, this data set is larger and more difficult since many of the images contain distracting digits at the sides of the digit of interest.
It contains 73,257 $32\times 32$ RGB images in the training set and 26,032 in the testing set.
Furthermore, it has a set of additional 531,131 images.
During the search we only use the training set, for the final training we make use of the additional examples as well.
We optimize the parameters with stochastic gradient descent with an initial learning rate of 0.025 and momentum of 0.9 for 40 epochs.
We decrease the learning rate by a factor of 0.5 after 5 epochs, after 10 epochs by another $2^{-7}$ and finally after 30 epochs by $2^{-11}$.
The weight decay was set to 0.0005.
We preprocess each image by subtracting each channel's mean and divide it by the standard deviation.

\subsubsection{CIFAR-10/CIFAR-100.}
The CIFAR-10 \cite{Krizhevsky2009} data set is a subset of the 80 million tiny images data set and its task is to identify one of ten different objects in $32\times 32$ RBG images.
The training split contains 50,000 examples, the testing split 10,000.
The CIFAR-100 data set is very similar to CIFAR-10 but with 10 times more classes.
We preprocess the image using global contrast normalization.
We use the same SGD optimizer as used for the SVHN.
However, we increased the number of epochs to 300.
We also use the same weight decay and reduce the learning rate in the same scheme but now after epochs 40, 80 and 240.
Additionally, we moderately augment the data by using random horizontal flips and random translation of up to 5 pixels.

We do not apply our search method on CIFAR-100.
Instead, we report the results obtained when training the best CIFAR-10 network architecture on CIFAR-100.

\begin{figure}
 \begin{center}
   \includegraphics[width=0.8\columnwidth]{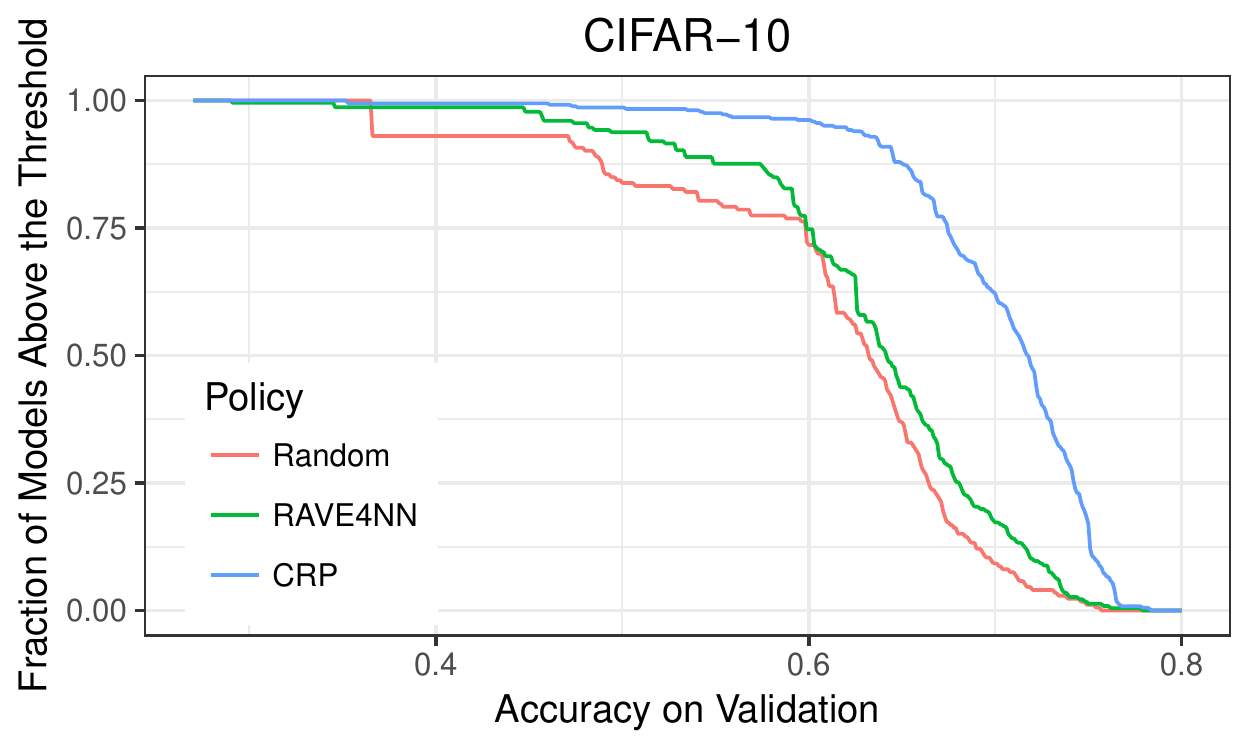}
 \caption{\label{fig:above-threshold}
  The fraction of models above a given threshold after the experiment on CIFAR-10.
  Not only the best model found by our methods is better than random but they also find better models on average.
  }
 \end{center}
\end{figure}
\begin{table*}
\caption{Comparison of different automatic neural network architecture search methods.
Results are reported in accuracy percent, duration in GPU days.
Our method is able to find competitive network architectures in just a single GPU day.
Results with asterisk (*) are obtained using the best architecture for CIFAR-10.
\label{tab:autonn-comparison}}
\begin{tabular*}{1\textwidth}{@{\extracolsep{\fill}}lrrrrr}
\hline\noalign{\smallskip}
Method & Duration (days) & MNIST & SVHN & CIFAR-10 & CIFAR-100\\
\noalign{\smallskip}
\hline
\noalign{\smallskip}
Neural Architecture Search \cite{Zoph2017} & 10,000 & - & - & 94.5 & - \\
Large-Scale Evolution \cite{Real2017} & 2,700  & - & - & 94.6 & 77.0\\
MetaQNN \cite{Baker2017} & 100 & 99.56 & 97.72 & 93.08 & 72.86* \\
RL with Network Transformation \cite{Cai2017} & 10 & - & - & 93.26 & - \\
DeepArchitect \cite{Negrinho2017} & 1.33 & - & - & 87.47 & - \\
\noalign{\smallskip}
\hline
\noalign{\smallskip}
Monte Carlo Planning (Random) & 1 & 99.48 & 96.73 & 84.97 & 61.09* \\
Monte Carlo Planning (\textsc{RAVE4NN}) & 1 & 99.69 & 96.89 & 90.23 & 66.84* \\
Monte Carlo Planning (\textsc{CRP}) & 1 & 99.51 & 97.24 & 91.20 & 69.75* \\
Monte Carlo Planning (\textsc{CRP}) & 5 & - & - & 93.55 & - \\
\hline
\end{tabular*}
\end{table*}
\begin{table}
\caption{Human proposed networks which use the same types of layers that we consider during our search.
\label{tab:human-comparison}}
\begin{tabular}{lrr}
\hline\noalign{\smallskip}
Method & CIFAR-10 & CIFAR-100\\
\noalign{\smallskip}
\hline
\noalign{\smallskip}
Maxout \cite{Goodfellow2013}& 90.62 & 61.43\\
Network in Network \cite{Lin2013}& 91.19 & 64.32\\
FitNet \cite{Romero2014}& 91.61 & 64.96\\
HighWay Network \cite{Srivastava2015}& 92.28 & -\\
All-CNN \cite{Springenberg2014}& 92.75 & 66.29\\
VGG \cite{Simonyan2014}& 92.75 & -\\
\hline
\end{tabular}
\end{table}
\subsection{Results.}

We present the search progress of the different policies for the data sets MNIST, SVHN and CIFAR-10 in Figure \ref{fig:search-progress}.
We report the validation performance of the best found model (solid lines) as well as the mean validation performance of the top five models (dashed lines).
CRP quickly finds good performing network architectures, clearly outperforming the random policy.
In many cases the top five performance is even able to outperform our proposed RAVE4NN policy.
The reason is that our reward prediction based on the current network structure allows us to learn good combinations of layers.
We see that this methods tends to start networks often with similar layer combinations.
Many of the networks for the CIFAR-10 data set start with a convolutional layer with kernel size $3\times 3$, followed by a pooling layer.
In many cases the next two layers are convolutional before adding another pooling layer.

The RAVE4NN policy needs some more time to converge against good network architectures because it has less information about the previous layers when deciding the next layer.
However, it is still outperforming the random policy and given enough time can catch up with the CRP policy.

Figure \ref{fig:above-threshold} gives an insight whether our proposed policies are providing indeed some useful improvements over the random policy.
It presents what fraction of neural networks achieves a validation accuracy above a threshold and provides insight into the policies' regret.
Both, RAVE4NN and CRP, are better than the random policy and are able to detect bad architectures and focus on more promising ones.
The results on the other data sets look similar and are moved to the supplementary material due to space constraints.

The network architecture with highest validation score is selected and retrained on the full training data set according to the description in Section \ref{sub:post_processing}.
We compare the obtained results on the testing data set against different state-of-the-art network architecture search methods in Table \ref{tab:autonn-comparison}.
We select a search budget of just a single GPU day to demonstrate that our method can be used also in case of limited hardware.
Thus, the duration is sometimes orders of magnitudes smaller than the one used by the competitor methods.
We did not reconduct the experiments for the competitor methods but report the results as in the corresponding papers.

DeepArchitect is the competitor method with the shortest searching time, approximately the time we gave our method.
However, both RAVE4NN and CRP are clearly outperforming its performance.
The test accuracy for DeepArchitect was not reported in the paper but the authors made this information available on their website \cite{Negrinho2017}.

The method by Cai et al. \cite{Cai2017} is the method with shortest running time that is outperforming our results.
However, their method does not start its search from scratch but rather start with a network of depth eight.
It is well structured, containing pooling, convolutional and fully connected layers.
While the authors claim that the network has a poor performance (73.1\% validation accuracy), we want to highlight that this is due to few parameters.
Since their reinforcement approach aims with its actions to add more filters for each layer, this is actually a very good starting point.
Thus, it is not surprising that our method which starts from scratch provides slightly worse results.

We consider MetaQNN \cite{Baker2017} to be our closest competitor.
This method is learning network architectures from scratch and we designed our action space to match exactly their setup.
They report a test accuracy of 93.08\% within 100 days for CIFAR-10 which is two orders of magnitudes more time than we needed for 91.2\%.
Their top five architectures achieve test accuracies from 88.37\% to 93.08\%.
The mean test accuracy of the top five architectures is 90.91\%, the standard deviation 1.68\%.
In comparison, our top five architectures for CRP achieve test accuracies between 89.16\% and 92.07\%, a mean of 90.80\% and a standard deviation of 1.10\%.
Thus, our method's best solution is not as good as the one by MetaQNN but the top networks provide consistent good test accuracies.
For this reason, an ensemble of the top five architectures improves the accuracy to 92.49\% while Baker et al. report a loss in accuracy to 92.68\% for MetaQNN \cite{Baker2017}.

Neural Architecture Search \cite{Zoph2017} and Large-Scale Evolution \cite{Real2017} achieve remarkable results.
However, the computation time is beyond anything most researchers and practitioners can afford.

An important question to answer is how good are the competitor methods when given less search time.
Real et al. report a test accuracy on CIFAR-10 of less than 30\% after one GPU day \cite{Real2017}.
Similarly, the Neural Architecture Search needs more than 800 model evaluations (much more than a GPU day) to show an improvement over a random policy \cite{Zoph2017}.
MetaQNN uses more than 50\% of its search time with random exploration \cite{Baker2017}.
This means, their results after more than 50 GPU days is as good as a random policy.
However, if we adapt their $\epsilon$ schedule to consider a maximum running time of one day instead of one hundred, only about 27 different networks will be evaluated.
15 of these network architectures are chosen completely at random, only for six architectures the actions are chosen according to the policy with a probability of more than 60\%.
Thus, it is very unlikely that this method finds a useful network in a single GPU day.
Cai et al. report an accuracy of about 87\% for their reinforcement learning method after the first half of networks was sampled \cite{Cai2017}.
In conclusion, our methods are indeed outperforming its competitors given a time budget of a single GPU day.

We conducted another experiment on CIFAR-10 giving our search method a five times higher time budget to see whether we can achieve equal performance if more time is invested.
During our one day experiment we noticed that our search algorithm converges against a policy that mainly chooses network architectures of depth seven.
The reason for this is that it learns to choose many pooling layers such that the representation size is reduced quickly.
In order to focus on deeper networks, we applied following changes.
First, pooling layers are constrained to be selected after at least three convolution layers.
Second, we slowly increase the minimum number of layers during the search.
Finally, we selected the network architecture with highest validation score and more than twelve layers.
This network was trained according to the previously described protocol on the full training data set and achieves an accuracy of 93.55\%.
Thus, we achieve very similar performance to the state-of-the-art in a fraction of the search time.

Further results and the found network architectures are reported in the supplementary material.

\section{Conclusions.}
We addressed the challenge of automating the neural network architecture engineering task in order to make deep learning accessible to a broader audience.
In particular, we were aiming at speeding up this procedure in order to make true end-to-end learning feasible for everyone.
As a solution, we proposed the use of Monte Carlo planning in combination with two different UCT derivations to search for network architectures.
To speed up the search, we adapted the UCT algorithm in order to share information between similar network architectures and made use of the Net2Net knowledge transfer.
In an empirical study we demonstrated that this method is able to find competitive networks for MNIST, SVHN and CIFAR-10 in just a single GPU day.
Extending the search time to five days, we can outperform the existing automated approaches and human proposed architectures which use the same types of layers.

\bibliographystyle{myplain}
\bibliography{myrefs}
\end{document}